\documentclass{article}

\usepackage[preprint]{neurips_2021}

\usepackage{float}

\usepackage{wrapfig}

\usepackage[utf8]{inputenc} 
\usepackage[T1]{fontenc}    
\usepackage{hyperref}       
\usepackage{url}            
\usepackage{booktabs}       
\usepackage{amsfonts}       
\usepackage{nicefrac}       
\usepackage{microtype}      
\usepackage{graphicx}
\usepackage{subcaption}
\usepackage{xcolor}         

\hypersetup{
    colorlinks = true,
    linkbordercolor = {white},
    citecolor={olive}
}

\usepackage{amsmath}
\usepackage{amsthm}
\usepackage{amssymb}
\usepackage{multirow}
\usepackage[ruled,vlined]{algorithm2e}
\newtheorem{theorem}{Theorem}[section]
\newtheorem{proposition}[theorem]{Proposition}
\newtheorem{lemma}[theorem]{Lemma}

\title{\Large Integrating Auxiliary Information in Self-supervised Learning}
\vspace{-5mm}
\author{%
  Yao-Hung Hubert Tsai$^{*}$, Tianqin Li\thanks{equal contribution}\,\,\,, Weixin Liu, Peiyuan Liao \\ {\bf Ruslan Salakhutdinov, Louis-Philippe Morency} \\
  Carnegie Mellon University
}

\begin{document}

\maketitle
\vspace{-6mm}
\begin{abstract}
   This paper presents to integrate the auxiliary information (e.g., additional attributes for data such as the hashtags for Instagram images) in the self-supervised learning process. We first observe that the auxiliary information may bring us useful information about data structures: for instance, the Instagram images with the same hashtags can be semantically similar. Hence, to leverage the structural information from the auxiliary information, we present to construct data clusters according to the auxiliary information. Then, we introduce the Clustering InfoNCE (Cl-InfoNCE) objective that learns similar representations for augmented variants of data from the same cluster and dissimilar representations for data from different clusters. Our approach contributes as follows: 1) Comparing to conventional self-supervised representations, the auxiliary-information-infused self-supervised representations bring the performance closer to the supervised representations; 2) The presented Cl-InfoNCE can also work with unsupervised constructed clusters (e.g., k-means clusters) and outperform strong clustering-based self-supervised learning approaches, such as the Prototypical Contrastive Learning (PCL) method; 3) We show that Cl-InfoNCE may be a better approach to leverage the data clustering information, by comparing it to the baseline approach - learning to predict the clustering assignments with cross-entropy loss. For analysis, we connect the goodness of the learned representations with the statistical relationships: i) the mutual information between the labels and the clusters and ii) the conditional entropy of the clusters given the labels.  
\end{abstract}

\vspace{-3mm}

\vspace{-1mm}
\section{Introduction}
\label{sec:intro}

Self-supervised learning  (SSL) considers the learning objectives that use data's self-information but not labels, where the labels are often expensive to collect. As a result, SSL empowers us to leverage a large amount of unlabeled data to learn good representations, and its applications span computer vision~\citep{chen2020simple,he2020momentum}, natural language processing~\citep{peters2018deep,devlin2018bert} and speech processing~\citep{schneider2019wav2vec,baevski2020wav2vec}. In addition to labels, we may sometimes access additional sources as auxiliary information for data, such as additional attributes information or data hierarchy information. The auxiliary information often naturally comes with the data, and hence it is cheaper to collect than labels. For example, Instagram images contain a mass amount of hashtags as additional attributes information. Nonetheless, the auxiliary information is often noisy. Hence, it raises a research challenge of effectively leveraging useful information from the auxiliary information in the SSL process.

\begin{figure}[t!]
\vspace{-5mm}
\begin{center}
\includegraphics[width=0.95\textwidth]{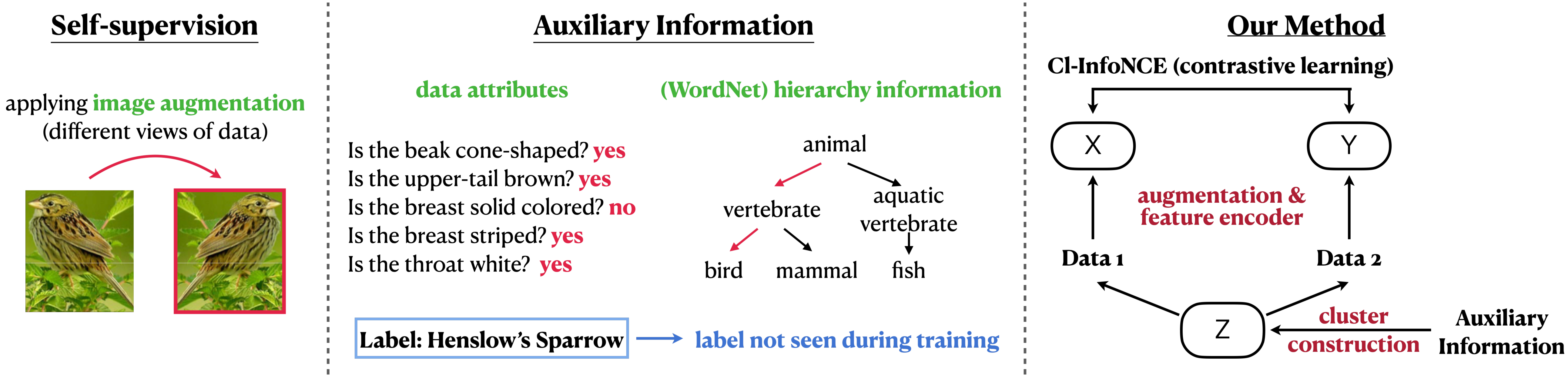}
\end{center}
\vspace{-4mm}
\caption{\small  {\bf Left: Self-supervision.} Self-supervised learning (SSL) uses self-supervision (the supervision from the data itself) for learning representations. An example of self-supervision is the augmented variant of the original data. 
{\bf Middle: Auxiliary Information.} This paper aims to integrate the auxiliary information into SSL. We consider two types of auxiliary information: data attributes and (WordNet) hierarchy information. In our example, the data attributes are binary indicators, and the hierarchy information is the hierarchy information for the label. {\bf Right: Our Method.} We first construct data clusters according to auxiliary information. We argue the formed clusters can provide valuable structural information of data for learning better representations. Second, we present the clustering InfoNCE (Cl-InfoNCE) objective to leverage the constructed clusters.}
\label{fig:illus}
\vspace{-4mm}
\end{figure}

We argue that a form of the valuable information provided by the auxiliary information is its implied clustering information of data. For example, we can expect an Instagram image to be semantically more similar to the image with the same hashtags than the image with different hashtags. Hence, our first step for leveraging the auxiliary information in SSL is to construct auxiliary-information-determined clusters. Specifically, we build data clusters such that the data from the same cluster have similar auxiliary information, such as having the same data attributes or belonging to the same data hierarchy. Then, our second step is to minimize the intra-cluster difference for the self-supervised learned representations. Particularly, we present the clustering InfoNCE (Cl-InfoNCE) objective to learn similar representations for augmented variants of data within the same cluster and dissimilar representations for data from different clusters. To conclude, the presented two-step approach leverages the structural information from the auxiliary information, then integrating the structural information into the SSL process. See Figure~\ref{fig:illus} for an overview of the paper.

We highlight several properties of our approach. 
First, we characterize the goodness of the Cl-InfoNCE-learned representations via the statistical relationships between the constructed clusters and the downstream labels. A resulting implication is that we can expect better downstream performance for our auxiliary-information-infused self-supervised representations when having i) higher mutual information between the labels and the auxiliary-information-determined clusters and ii) lower conditional entropy of the clusters given the labels. Second, Cl-InfoNCE generalizes recent contrastive learning objectives by changing the way to construct the clusters. In particular, when each cluster contains only one data, Cl-InfoNCE specializes in conventional self-supervised contrastive objective (e.g., the InfoNCE objective~\citep{oord2018representation}). When the clusters are labels, Cl-InfoNCE specializes in supervised contrastive objective (e.g., the objective considered by~\citet{khosla2020supervised}). The generalization implies that our approach (auxiliary-information-determined clusters + Cl-InfoNCE) works between conventional self-supervised and supervised representation learning.
Third, Cl-InfoNCE is a computationally efficient method as it can scale up even with many clusters. The reason is that Cl-InfoNCE is a contrastive-based approach, which is non-parametric. Particularly, the number of the parameters in Cl-InfoNCE is independent of the number of clusters. 

We conduct experiments on learning visual representations using UT-zappos50K~\citep{yu2014fine}, CUB-200-2011~\citep{wah2011caltech}, Wider Attribute~\citep{li2016human} and ImageNet-100~\citep{russakovsky2015imagenet} datasets. For the first set of experiments, we focus on the analysis of Cl-InfoNCE to study how well it works with unsupervised constructed clusters (K-means clusters). We find it achieves better performance comparing to the clustering-based self-supervised learning approaches, such as the Prototypical Contrastive Learning (PCL)~\citep{li2020prototypical} method. The result suggests that the K-means method + Cl-InfoNCE can be a strong baseline for the conventional self-supervised learning setting. For the second set of experiments, we like to see how much improvement can the auxiliary information bring to us. We consider the discrete attributes and the WordNet hierarchy information~\citep{miller1995wordnet} as the auxiliary information. We show that the auxiliary-information-infused self-supervised representations, compared to conventional self-supervised representation, have a much better performance on downstream tasks. We also find that Cl-InfoNCE has a better performance than the baseline - predicting the clustering assignments with cross-entropy loss.

\vspace{-1mm}
\section{Related Work}
\label{sec:rela}

\vspace{-1mm}
\paragraph{Self-supervised Learning.} Self-supervised learning (SSL) defines a pretext task as a pre-training step and uses the pre-trained features for a wide range of downstream tasks, such as object detection and segmentation in Computer Vision~\citep{chen2020simple,he2020momentum}, question answering, and language understanding in Natual Language Processing~\citep{peters2018deep,devlin2018bert} and automatic speech recognition in Speech Processing~\citep{schneider2019wav2vec,baevski2020wav2vec}. In this paper, we focus on discussing two types of pretext tasks: clustering approaches~\citep{caron2018deep,caron2020unsupervised} and contrastive approaches~\citep{chen2020simple,he2020momentum}.

On the one hand, the clustering approaches jointly learn the networks' parameters and the cluster assignments of the resulting features. The cluster assignments are obtained through unsupervised clustering methods such as k-means~\citep{caron2018deep}, the optimal transportation algorithms such as Sinkhorn algorithm~\citep{caron2020unsupervised}, etc. It is worth noting that the clustering approaches enforce consistency between cluster assignments for different augmentations of the same data. On the other hand, the contrastive approaches learn similar representations for augmented variants of a data and dissimilar representations for different data. The objectives considered for contrastive approaches are the InfoNCE objective~\citep{oord2018representation,chen2020simple,he2020momentum}, Wasserstein Predictive Coding~\citep{ozair2019wasserstein}, Relative Predictive Coding~\citep{tsai2021self}, etc. Both the clustering and the contrastive approaches aim to learn representations that are invariant to data augmentations. 

There is another line of work combining clustering and contrastive approaches, such as HUBERT~\citep{hsu2020hubert}, Prototypical Contrastive Learning~\citep{li2020prototypical} and Wav2Vec~\citep{schneider2019wav2vec,baevski2020wav2vec}. They first construct (unsupervised) clusters from the data. Then, they perform a contrastive approach to learn similar representations for the data within the same cluster. Our approach relates to these work with two differences: 1) we construct the clusters from the auxiliary information; and 2) we present Cl-InfoNCE as a new contrastive approach and characterize the goodness for the resulting representations.

\vspace{-1mm}
\paragraph{Learning to Predict Auxiliary Information.} Our study also relates to work on learning to predict weak labels~\citep{sun2017revisiting,mahajan2018exploring,wen2018disjoint,radford2021learning}. The weak labels can be hashtags for Instagram images~\citep{mahajan2018exploring}, metadata such as identity and nationality for a person~\citep{wen2018disjoint} or corresponding textual descriptions for an image~\citep{radford2021learning}. Compared to labels, the weak labels are noisy but require much less manual annotation work. This line of work shows that the network learned by weakly supervised pre-training tasks can generalize well to various downstream tasks, including object detection and segmentation, cross-modality matching, and video action recognition. The main difference between this line of work and ours is that our approach does not consider a prediction objective but a contrastive learning objective (i.e., the Cl-InfoNCE objective).

\vspace{-2mm}
\section{Method}
\label{sec:method}

We present a two-step approach to leverage the structural information from the auxiliary information and then integrate this structural information into the self-supervised learning process. The first step (Section~\ref{subsec:cluster_construct}) clusters data according to auxiliary information. And we consider discrete attributes and data hierarchy as the auxiliary information.
The second step (Section~\ref{subsec:cl-infonce}) presents the clustering InfoNCE (Cl-InfoNCE) objective, a contrastive-learning-based approach, to leverage the constructed clusters. 
Last, in Section~\ref{subsec:impli_inves}, we discuss the implications and provide the investigations for our approach.
For notations, we use the upper case (e.g., $X$) letter to denote the random variable and the lower case (e.g., $x$) to denote the outcome from the random variable. 

\vspace{-1mm}
\subsection{Cluster Construction for Discrete Attributes and Data Hierarchy Information}
\label{subsec:cluster_construct}

\begin{figure}[t!]
\vspace{-4mm}
\begin{center}
\includegraphics[width=1.0\textwidth]{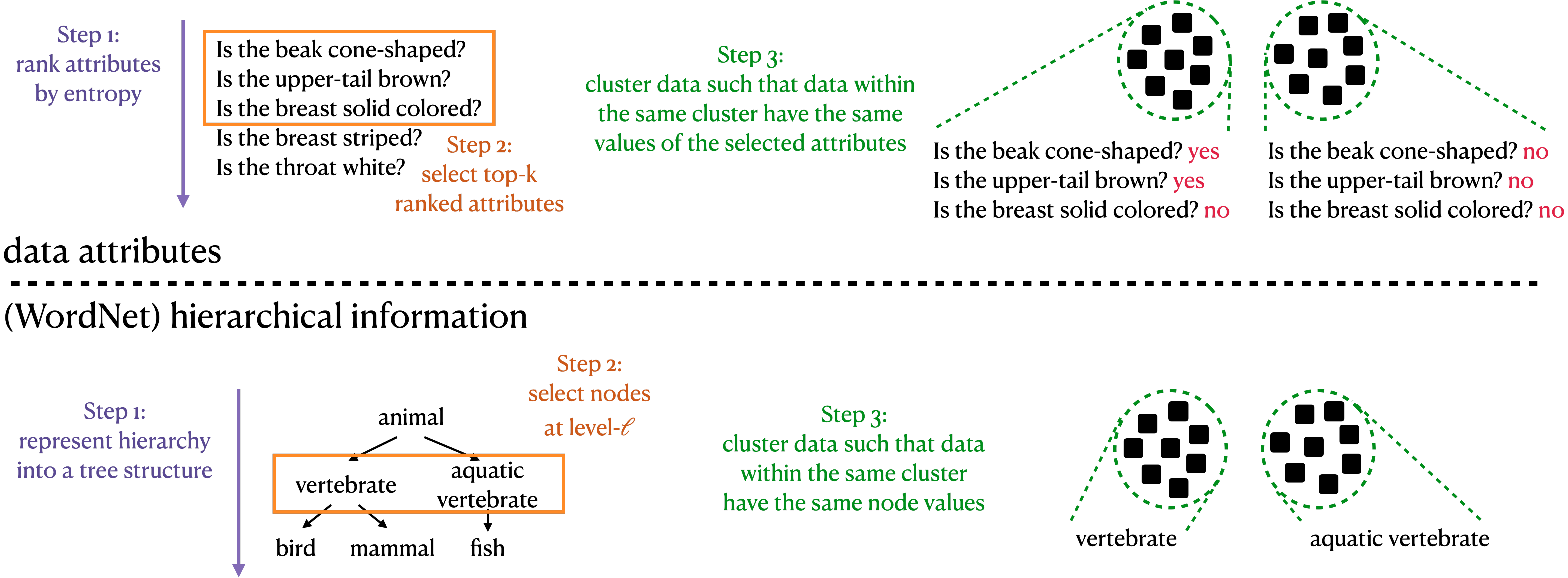}
\end{center}
\vspace{-4mm}
\caption{\small  Cluster construction according to auxiliary information. We consider data attributes and (WordNet) hierarchical information as auxiliary information.}
\label{fig:cluster_construction}
\vspace{-3mm}
\end{figure}

This sub-Section discusses how we construct data clusters according to auxiliary information. And in this paper, we consider the data attributes and data hierarchy information as the auxiliary information. Note that the cluster constructions may differ with different types of auxiliary information. Below, we present our specific ways to determine data clusters according to our selected types of auxiliary information. We focus on providing overviews of our method, and more details can be found in our released code\footnote{Anonymous Link.}. We provide the illustration in Figure~\ref{fig:cluster_construction}.

\vspace{-1mm}
\paragraph{Clustering according to Discrete Attributes.} We consider the discrete attributes as the first type of auxiliary information. An example of such auxiliary information is binary indicators of attributes, such as ``short/long hair'', ``with/without sunglasses'' or ``short/long sleeves'', for human photos. We construct the clusters such that data within each cluster will have the same values for a set of attributes. In our running example, if picking the set of attributes being hair and sunglasses, the human photos having both the ``long hair'' and ``with sunglasses'' will form a cluster. Then, how we determine the set of attributes? First, we rank each attribute according to its entropy in the dataset. Note that if an attribute has high entropy, it means this attribute is distributed diversely. Then, we select the attributes with top-$k$ highest entropy, where $k$ is a hyper-parameter.

\paragraph{Clustering according to Hierarchy Information.} As the second type of auxiliary information, we consider hierarchy information - more specifically, the WordNet hierarchy~\citep{miller1995wordnet}. The WordNet hierarchy describes the hierarchy information for data labels. For instance, assuming ``human'' and ``mouse'' as the labels, WordNet hierarchy suggests 1) ``mammal'' is the parent of ``human'' and ``mouse''; and 2) ``vertebrate'' is the parent of ``mammal''. In this running example, ``mammal'' and ``vertebrate'' can be seen as the coarse labels of data, and we construct the clusters such that data within each cluster will have the same coarse label. Then, how we choose the coarse labels? We first represent the WordNet hierarchy into a tree structure (each children node has only one parent node). 
Then, we choose the coarse labels to be the nodes in the level $l$ in the WordNet tree hierarchy (the root node is level $1$). $l$ is a hyper-parameter.

\vspace{-1mm}
\subsection{Clustering InfoNCE Objective}
\label{subsec:cl-infonce}

So far, we see how we determine the data clusters from discrete data attributes or data hierarchy information (as the auxiliary information). Now, we shall show how we integrate this clustering information into the self-supervised learning process. We note that most of the self-supervised learning approaches present to learn representations invariant to data augmentations~\citep{chen2020simple,caron2020unsupervised}. And on this basis, we present to learn representations that will also be similar for data with the same cluster assignment. To this end, we introduce the clustering InfoNCE (Cl-InfoNCE) objective, which is inspired by the InfoNCE objective~\citep{oord2018representation} (which is widely used in conventional self-supervised representation learning). For a better presentation flow, we leave the discussion of InfoNCE later (in Section~\ref{subsec:impli_inves}) but do not present it as a technical background first. We use the alphabets $X$ and $Y$ to denote the representations from augmented data:
\begin{equation*}
\resizebox{\hsize}{!}{$X={\rm Feature\_Encoder\Big(Augmentation\_1\big(Data\_1\big)\Big)}\,\,{\rm and}\,\,Y={\rm Feature\_Encoder\Big(Augmentation\_2\big(Data\_2\big)\Big)}$},
\end{equation*}
and the alphabet $Z$ to denote the constructed clusters. 
Then, we formulate Cl-InfoNCE as
\vspace{1mm}
\begin{proposition}[Clustering-based InfoNCE (Cl-InfoNCE)]
\vspace{-1mm}
\begin{equation}
{\rm Cl-InfoNCE}:=\underset{f}{\rm sup}\,\,\mathbb{E}_{(x_i, y_i)\sim {\mathbb{E}_{z\sim P_Z}\big[P_{X|z}P_{Y|z}\big]}^{\otimes n}}\Big[ {\rm log}\,\frac{e^{f(x_i, y_i)}}{\frac{1}{n}\sum_{j=1}^n e^{f(x_i, y_j)}}\Big],
\label{eq:cl_infonce}
\end{equation}
\end{proposition}
\vspace{-1mm}
where $f(x,y)$ is any function that returns a scalar from the input $(x,y)$. As suggested by prior work~\citep{chen2020simple,he2020momentum}, we choose $f(x,y) = {\rm cosine}\big(g(x),g(y)\big) / \tau$ to be the cosine similarity between non-linear projected $g(x)$ and $g(y)$. $g(\cdot)$ is a neural network (also known as the projection head~\citep{chen2020simple,he2020momentum}) and $\tau$ is the temperature hyper-parameter. $\{(x_i, y_i)\}_{i=1}^n$ are $n$ independent copies of $(x,y)\sim \mathbb{E}_{z\sim P_Z}\big[P_{X|z}P_{Y|z}\big]$, where it first samples a cluster $z \sim P_Z$ and then samples $(x,y)$ pair with $x \sim P_{X|z}$ and $y \sim P_{Y|z}$. Furthermore, we call $(x_i,y_i)$ as the positively-paired data ($x_i$ and $y_i$ have the same cluster assignment) and $(x_i,y_j)$ ($i\neq j$) as the negatively-paired data ($x_i$ and $y_j$ have independent cluster assignment). Note that, in practice, the expectation in eq.~\eqref{eq:cl_infonce} is replaced by the empirical mean of a batch of samples.

Our objective is learning the representations $X$ and $Y$ (by updating the parameters in the feature encoder) to maximize Cl-InfoNCE. At a colloquial level, the maximization pulls towards the representations of the augmented data within the same cluster and push away the representations of the augmented data from different clusters. Theoretically, we present the following:
\begin{theorem}[informal, Cl-InfoNCE maximization learns to include the clustering information]
\vspace{-1.5mm}
\begin{equation}
\begin{split}
    & {\rm Cl-InfoNCE} \leq D_{\rm KL}\,\Big( \mathbb{E}_{P_Z}\big[P_{X|Z}P_{Y|Z}\big] \,\|\, P_{X}P_{Y} \Big) \leq H(Z) \\
{\rm and}\,\,& {\rm the\,\,equality\,\,holds\,\,only\,\,when \,\,} H(Z|X) = H(Z|Y) = 0,
\end{split}
\label{eq:max_resulting_repre}
\end{equation}
\vspace{-3mm}
\label{theo:max_resulting_repre}
\end{theorem}
where $H(Z)$ is the entropy of $Z$ and $H(Z|X)$ (or $H(Z|Y)$) are the conditional entropy of $Z$ given $X$ (or $Y$). Please find detailed derivations and proofs in Appendix. 

The theorem suggests that Cl-InfoNCE has an upper bound $D_{\rm KL}\,\Big( \mathbb{E}_{P_Z}\big[P_{X|Z}P_{Y|Z}\big] \,\|\, P_{X}P_{Y} \Big)$, which measures the distribution divergence between the product of clustering-conditional marginal distributions (i.e., $\mathbb{E}_{P_Z}\big[P_{X|Z}P_{Y|Z}\big]$) and the product of marginal distributions (i.e., $P_{X}P_{Y}$). We give an intuition for $D_{\rm KL}\,\Big( \mathbb{E}_{P_Z}\big[P_{X|Z}P_{Y|Z}\big] \,\|\, P_{X}P_{Y} \Big)$: if $D_{\rm KL}\,\Big( \mathbb{E}_{P_Z}\big[P_{X|Z}P_{Y|Z}\big] \,\|\, P_{X}P_{Y} \Big)$ is high, then we can easily tell whether $(x,y)$ have the same cluster assignment or not. The theorem also suggests that maximizing Cl-InfoNCE results in the representations $X$ and $Y$ including the clustering information $Z$ ($\because H(Z|X) = H(Z|Y) = 0$). 

\vspace{-1mm}
\subsection{Implications and Investigations}
\label{subsec:impli_inves}


\paragraph{Goodness of the Learned Representations.} 
In Theorem~\ref{theo:max_resulting_repre}, we show that maximizing Cl-InfoNCE learns the representations ($X$ and $Y$) to include the clustering ($Z$) information. Therefore, to characterize how good is the learned representations by maximizing Cl-InfoNCE, we can instead study the relations between $Z$ and the downstream labels (denoting by $T$). In particular, we can use information-theoretical metrics such as the mutual information $I(Z;T)$ and the conditional entropy $H(Z|T)$ to characterize the goodness of the learned representations. $I(Z;T)$ measures how relevant the clusters and the labels, and $H(Z|T)$ measures how much redundant information in the clusters that are irrelevant to the labels.  
For instance, we can expect good downstream performance for our auxiliary-information-infused representations when having high mutual information and low conditional entropy between the auxiliary-information-determined clusters and the labels.

\vspace{-2mm}
\paragraph{Generalization of Recent Self-supervised and Supervised Contrastive Approaches.} Cl-InfoNCE (eq.~\eqref{eq:cl_infonce}) serves as an objective that generalizes to different levels of supervision according to how we construct the clusters ($Z$). When $Z=$ instance id (i.e., each cluster only contains an instance), $\mathbb{E}_{P_Z}\big[P_{X|Z}P_{Y|Z}\big]$ specializes to $P_{XY}$ and Cl-InfoNCE specializes to the InfoNCE objective~\citep{oord2018representation}, which aims to learn similar representations for augmented variants of the same data and dissimilar representations for different data. InfoNCE is the most popular used self-supervised contrastive learning objective~\citep{chen2020simple,he2020momentum,tsai2021multiview}. When $Z=$ downstream labels, Cl-InfoNCE specializes to the objective described in {\em Supervised Contrastive Learning}~\citep{khosla2020supervised}, which aims to learn similar representations for data that are from the same downstream labels and vice versa. In our paper, the clusters $Z$ are determined by the auxiliary information, and we aim to learn similar representations for data sharing the same auxiliary information and vice versa. This process can be understood as weakly supervised contrastive learning. To conclude, Cl-InfoNCE is a clustering-based contrastive learning objective. By differing its cluster construction, Cl-InfoNCE interpolates among unsupervised, weakly supervised, and supervised representation learning.

\vspace{-2mm}
\paragraph{Advantages over Learning to Predict the Clusters Assignments.} An alternative way to leverage the data clustering information is learning to predict the cluster assignment ($Z$) from the representations ($X$ and $Y$). An example is learning to predict the hashtags for Instagram images~\citep{mahajan2018exploring}, where the author shows that this prediction process serves as a good pre-training step. Nonetheless, comparing to our presented Cl-InfoNCE objective, learning to predict the cluster assignment requires building an additional classifier between the representations and the cluster. It will be non-ideal and inefficient to optimize this classifier when having a large number of clusters. The reason is that the number of the classifier's parameters is proportional to the number of clusters. An example is that, when $Z=$ instance id, the number of the clusters will be the total number of data, which can be billions. Learning to predict the clustering assignment may work poorly under this case, while InfoNCE (Cl-InfoNCE when $Z=$ instance id) can reach a good performance~\citep{chen2020simple}. Last, the most used objective for learning to predict the clusters is the cross-entropy loss. And evidences~\citep{khosla2020supervised} show that, compared to the cross-entropy loss, the contrastive objective (e.g., our presented Cl-InfoNCE) is more robust to natural corruptions of data and stable to hyper-parameters and optimizers settings.
\section{Experiments}
In the beginning, we discuss the datasets used in the paper in Section~\ref{subsec:datasets}. We consider either discrete attributes or data hierarchy information as auxiliary information for data. Then, in Section~\ref{subsec:method}, we explain the methodology that will be used in the experiments. In Section~\ref{subsec:exp_without_auxi}, we present the first set of the experiments, which focuses on studying the presented Cl-InfoNCE objective (see Section~\ref{subsec:cl-infonce}) under conventional self-supervised setting. To this end, we consider unsupervised constructed clusters (e.g., k-means) along with Cl-InfoNCE. 
And we compare Cl-InfoNCE with other clustering-based self-supervised approaches. 
In Section~\ref{subsec:exp_with_auxi_attr} and~\ref{subsec:exp_with_auxi_hier}, we further present experiments under the scenario when auxiliary information is available. We compare our method with the baseline approach - learning to predict the clustering assignment with cross-entropy loss. We also compare with conventional self-supervised representations and supervised representations.

\subsection{Datasets}
\label{subsec:datasets}

We consider the following datasets. {\bf UT-zappos50K}~\citep{yu2014fine}: It contains $50,025$ shoes images along with $7$ discrete attributes as auxiliary information. Each attribute follows a binomial distribution, and we convert each attribute into a set of Bernoulli attributes, resulting in a total of $126$ binary attributes. There are $21$ shoe categories.
{\bf Wider Attribute}~\citep{li2016human}: It contains $13,789$ images, and there are several bounding boxes in an image. The attributes are annotated per bounding box. We perform OR operation on attributes from different bounding boxes in an image, resulting in $14$ binary attributes per image as the auxiliary information. There are $30$ scene categories. 
{\bf CUB-200-2011}~\citep{wah2011caltech}: It contains $11,788$ bird images with $312$ binary attributes as the auxiliary information. There are $200$ bird species. {\bf ImageNet-100}~\citep{russakovsky2015imagenet}: It is a subset of the ImageNet-1k object recognition dataset~\citep{russakovsky2015imagenet}, where we select $100$ categories out of $1,000$, resulting in around $0.12$ million images. We consider WordNet hierarchy information as the auxiliary information.

\subsection{Methodology}
\label{subsec:method}

Following~\citet{chen2020simple}, we conduct experiments on pre-training visual representations and then evaluating the learned representations using the linear evaluation protocol. In precise, after the pre-training stage, we fix the pre-trained feature encoder and then categorize test images by linear classification results. We select ResNet-50~\citep{he2016deep} as our feature encoder across all settings. Note that our goal is learning representations (i.e, $X$ and $Y$) for maximizing the Cl-InfoNCE objective (equation~\eqref{eq:cl_infonce}). Within Cl-InfoNCE, the positively-paired representations $(x, y^+) \sim \mathbb{E}_{z\sim P_Z}\big[P_{X|z}P_{Y|z}\big]$ are the learned representations from augmented images from the same cluster $z\sim P_Z$ and the negatively-paired representations $(x, y^-) \sim P_XP_Y$ are the representations from arbitrary two images. We leave the network designs, the optimizer choices, and more details for the datasets in Appendix.

\begin{wrapfigure}{r}{0.42\textwidth}
\centering
\vspace{-7mm}
\hspace{-6mm}
\includegraphics[width=0.4\textwidth]{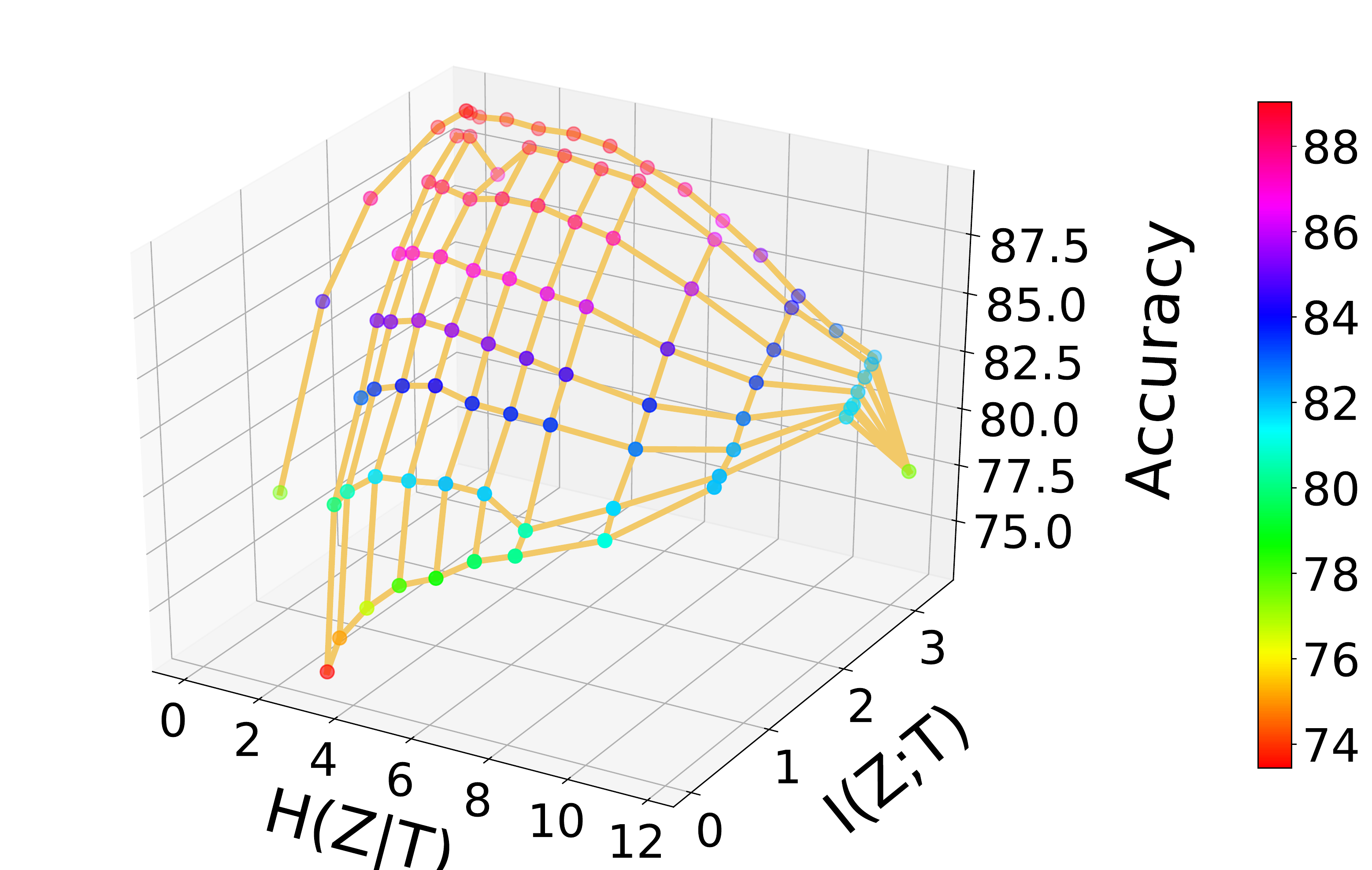}
\caption{\small $I(Z;T)$ represents how relevant the clusters and the labels; higher is better. $H(Z|T)$ represents the redundant information in the clusters for the labels; lower is better.}\label{fig:information-plane}
\vspace{-4mm}
\end{wrapfigure}
Before delving into the experiments, we like to recall that, in Section~\ref{subsec:impli_inves}, we discussed using the mutual information $I(Z;T)$ and the conditional entropy $H(Z|T)$ between the clusters ($Z$) and the labels ($T$) to characterize the goodness of Cl-InfoNCE's learned representations. To prove this concept, on UT-Zappos50K, we synthetically construct clusters for various $I(Z;T)$ and $H(Z|T)$ followed by applying Cl-InfoNCE. We present the results in the right figure.
Our empirical results are in accordance with the statements that the clusters with higher $I(Z;T)$ and lower $H(Z|T)$ will lead to higher downstream performance. In later experiments, we will also discuss these two information-theoretical metrics.

\subsection{Experiment I: K-means Clusters + Cl-InfoNCE}
\label{subsec:exp_without_auxi}

We study how Cl-InfoNCE can learn good self-supervised representations even without auxiliary information. To this end, we construct unsupervised clusters (e.g., k-means clusters on top of the learned representations) for Cl-InfoNCE. Similar to the EM algorithm, we iteratively perform the k-means clustering to determine the clusters for the representations, and then we adopt Cl-InfoNCE to leverage the k-means clusters to update the representations. We select the Prototypical Contrastive Learning (PCL)~\citep{li2020prototypical} as the baseline of the clustering-based self-supervised approach. In particular, PCL performs data log-likelihood maximization by assuming data are generated from isotropic Gaussians. It considers the MLE objective, where the author makes a connection with contrastive approaches~\citep{chen2020simple,he2020momentum}. The clusters in PCL are determined via MAP estimation. For the sake of the completeness of the experiments, we also include the non-clustering-based self-supervised approaches, including SimCLR~\citep{chen2020simple} and MoCo~\citep{he2020momentum}. Note that this set of experiments considers the conventional self-supervised setting, in which we can leverage the information neither from labels nor from auxiliary information. 

\begin{figure}[t!] 
\vspace{-5mm}
\centering
\hspace{-1mm}
\begin{minipage}{0.67\linewidth}
\scalebox{0.59}{
\begin{tabular}{c|c|c|c|c}
\toprule
\multirow{2}{*}{Method}              & \multicolumn{1}{c|}{\textbf{UT-Zappos50K}} & \multicolumn{1}{c|}{\textbf{Wider Attribute}} & \multicolumn{1}{c|}{\textbf{CUB-200-2011}} & \multicolumn{1}{c}{\textbf{ImageNet-100}} \\
                                     & Top-1 (Accuracy)                & Top-1 (Accuracy)             & Top-1 (Accuracy)  &   Top-1 (Accuracy)     \\ \midrule \midrule  \multicolumn{5}{c}{\it Non-clustering-based Self-supervised Approaches}  \\ \midrule \midrule
SimCLR~\citep{chen2020simple}       & 77.8$\pm$1.5 & 40.2$\pm$0.9       & 14.1$\pm$0.7      & 58.2$\pm$1.7  \\ [1mm]
MoCo~\citep{he2020momentum}      & 83.4$\pm$0.5               &  41.0$\pm$0.7              & 13.8$\pm$0.5       & 59.4$\pm$1.6     \\ \midrule \midrule \multicolumn{5}{c}{\it Clustering-based Self-supervised Approaches (\# of clusters = $1$K/ $1$K/ $1$K/ $2.5$K)} \\ \midrule \midrule
PCL~\citep{li2020prototypical}      & 82.4$\pm$0.5                & 41.0$\pm$0.4               & 14.4$\pm$0.5            & 68.9$\pm$0.7         \\ [1mm]
K-means + Cl-InfoNCE (ours)      & \textbf{84.5$\pm$0.4}               & \textbf{43.6$\pm$0.4}               & \textbf{17.6$\pm$0.2}            & \textbf{77.9$\pm$0.7}             \\ \bottomrule
\end{tabular}
}
\end{minipage}
\hspace{1mm}
\begin{minipage}{0.31\linewidth}
\includegraphics[width=\linewidth]{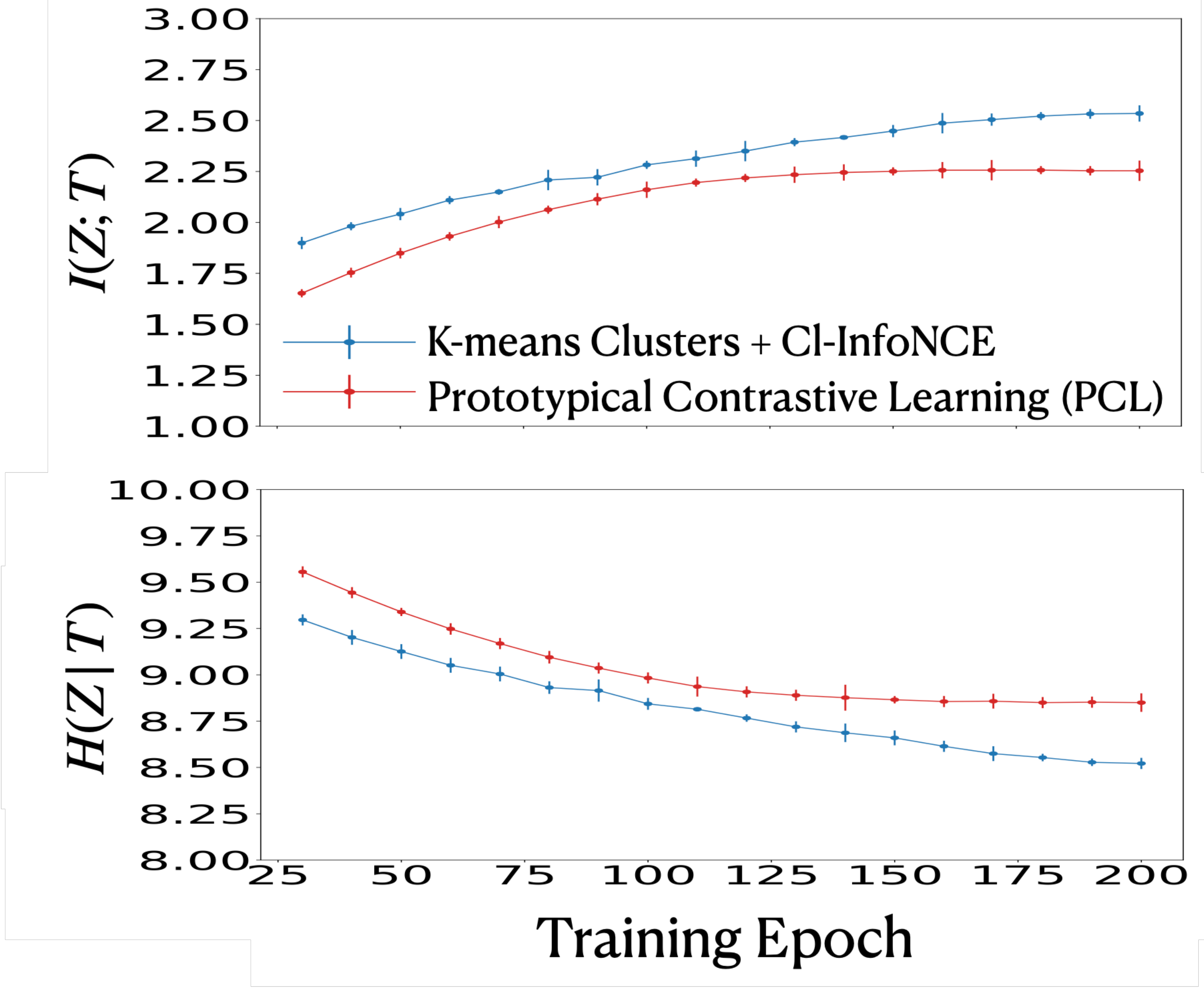}
\end{minipage}
\vspace{-2mm}
\caption{\small 
Experimental results under conventional self-supervised setting (pre-training using no label supervision and no auxiliary information). {\bf Left:} We compare our method (K-means clusters + Cl-InfoNCE) with self-supervised approaches that leverage and do not consider unsupervised clustering. The downstream performance is reported using the linear evaluation protocal~\citep{chen2020simple}. {\bf Right:} For our method and Prototypical Contrastive Learning (PCL), we plot the mutual information ($I(Z;T)$) and the conditional entropy ($H(Z|T)$) versus training epochs. $Z$ are the unsupervised clusters, and $T$ are the downstream labels.}
\vspace{-4mm}
\label{fig:unsupervised}
\end{figure}

\paragraph{Results.} We first look at the left table in Figure~\ref{fig:unsupervised}. We observe that, except for ImageNet-100, there is no obvious performance difference between the non-clustering-based (i.e., SimCLR and MoCo) and the clustering-based baseline (i.e., PCL). Since ImageNet-100 is a more complex dataset comparing to the other three datasets, we argue that, when performing self-supervised learning, discovering latent structures in data (via unsupervised clustering) may best benefit larger-sized datasets. Additionally, among all the approaches, our method reaches the best performance. The result suggests our method can be as competitive as other conventional self-supervised approaches. 

Next, we look at the right plot in Figure~\ref{fig:unsupervised}. We study the mutual information $I(Z;T)$ and the conditional entropy $H(Z|T)$ between the unsupervised constructed clusters $Z$ and the downstream labels $T$. We select our method and PCL, providing the plot of the two information-theoretical metrics versus the training epoch. We find that, as the number of training epochs increases, both methods can construct unsupervised clusters that are more relevant (higher $I(Z;T)$) and contain less redundant information (lower $H(Z|T)$) about the downstream label. This result suggests that the clustering-based self-supervised approaches are discovering the latent structures that are more useful for the downstream tasks. It is worth noting that our method consistently has higher $I(Z;T)$ and lower $H(Z|T)$ comparing to PCL.

\subsection{Experiment II: Data-Attributes-Determined Clusters + Cl-InfoNCE}
\label{subsec:exp_with_auxi_attr}

We like to understand how well Cl-InfoNCE can be combined with the auxiliary information. For this purpose, we select the data discrete attributes as the auxiliary information, construct the clusters ($Z$) using the discrete attributes (see Section~\ref{subsec:cluster_construct} and Figure~\ref{fig:cluster_construction}), and then adopt attributes-determined clusters for Cl-InfoNCE. Recall our construction of data-attributes-determined clusters: we select the attributes with top-$k$ highest entropy and then construct the clusters such that the data within a cluster will have the same values over the selected attributes. $k$ is the hyper-parameter. Note that our method considers a weakly supervised setting since the data attributes can be seen as the data's weak supervision. For the completeness of the experiments, we include the comparisons with the supervised ($Z=$ downstream labels $T$) and the conventional self-supervised ($Z=$ instance ID) setting for our method. We show in Section~\ref{subsec:impli_inves}, the supervised setting is equivalent to the Supervised Contrastive Learning objective~\citep{khosla2020supervised} and the conventional self-supervised setting is equivalent to SimCLR~\citep{chen2020simple}. We also include another baseline that leverages the data clustering information - learning to predict the clusters assignments using cross-entropy loss.

\begin{table}[t!]
\vspace{-5mm}
\centering
\scalebox{0.65}{
\begin{tabular}{c|cc|cc|cc}
\toprule
\multirow{2}{*}{Method (Contrastive Learning$^\dagger$ / Predictive Learning$^\ddagger$)}              & \multicolumn{2}{c|}{\textbf{UT-Zappos50K}} & \multicolumn{2}{c|}{\textbf{Wider Attribute}} & \multicolumn{2}{c}{\textbf{CUB-200-2011}} \\
                                     & Top-1 Acc.              & Top-5 Acc.              & Top-1 Acc.            & Top-5 Acc.            & Top-1 Acc.               & Top-5 Acc.               \\ \midrule \midrule \multicolumn{7}{c}{\it Supervised Representation Learning ($Z=$ downstream labels $T$)} \\ \midrule \midrule
$^\ddagger$Cross-Entropy Loss                        & 89.2$\pm$0.5               & 99.6$\pm$0.4               & 44.7$\pm$1.5             & 71.2$\pm$ 0.5           & 60.5$\pm$1.2                & 81.7$\pm$0.7                \\ [1mm]
$^\dagger$(Labels + Cl-InfoNCE) SupCon~\citep{khosla2020supervised}      & 89.0$\pm$0.4                & 99.4$\pm$ 0.3              & 49.9$\pm$0.8             & 76.2$\pm$0.2             & 59.9$\pm$0.7                & 78.8$\pm$ 0.3               \\ \midrule \midrule \multicolumn{7}{c}{\it Weakly Supervised Representation Learning ($Z=$ attributes-determined clusters)} \\ \midrule \midrule
$^\ddagger$Cross-Entropy Loss   & 82.7$\pm$0.7                & 99.04$\pm$0.3               & 39.4$\pm$0.6          & 68.6$\pm$0.2           & 17.5$\pm$1.0                & 46.0$\pm$0.8   \\ [1mm]
$^\dagger$Attributes-Determined Clusters + Cl-InfoNCE (ours)      & 84.6$\pm$0.4                & 99.1$\pm$0.2               & 45.5$\pm$0.2            & 75.4$\pm$0.2             & 20.6$\pm$ 0.5               & 47.0$\pm$0.5                \\ 
  \midrule \midrule \multicolumn{7}{c}{\it Self-supervised Representation Learning ($Z=$ instance id)} \\
 \midrule \midrule
 $^\dagger$MoCo~\citep{he2020momentum} & 83.4$\pm$0.2                & 99.1$\pm$0.3               & 41.03$\pm$0.7             & 74.0$\pm$0.4             & 13.8$\pm$0.7               & 36.5$\pm$0.5
                \\ [1mm]
    $^\dagger$(Instance-ID + Cl-InfoNCE) SimCLR~\citep{chen2020simple} & 77.8$\pm$1.0                & 97.9$\pm$0.8               & 40.2$\pm$0.9             & 73.0$\pm$0.3             & 14.1$\pm$ 0.7               & 35.2$\pm$0.6            \\\bottomrule
\end{tabular}
}
\vspace{1mm}
\caption{\small Experimental results under supervised (pre-training using label supervision), weakly supervised (pre-training using data attributes), and conventional self-supervised (pre-training using neither label supervision nor data attributes) setting. Each setting refers to a particular cluster ($Z$) construction. The methods presented in this table are either contrastive or predictive learning approaches. We report the best results for weakly supervised methods by tuning the hyper-parameter $k$ for attributes-determined clusters. 
}
\label{tbl:attributes}
\vspace{-3mm}
\end{table}
\begin{figure}[t!]
    \centering
       \vspace{-3mm} \includegraphics[width=\textwidth]{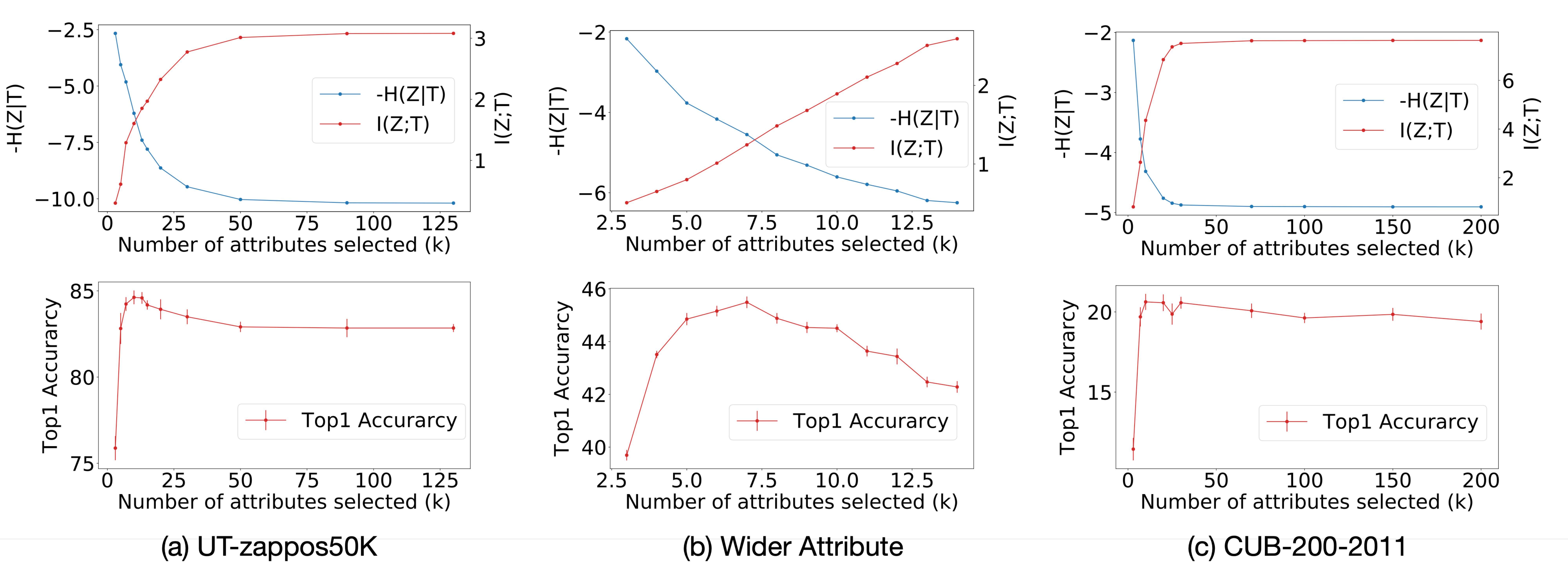}
        \vspace{-6mm}
        \caption{\small Experimental results for attributes-determined clusters + Cl-InfoNCE by tuning the hyper-parameter $k$ when constructing the clusters. Note that we select attributes with top-$k$ highest entropy, and we construct the clusters such that the data within a cluster would have the same values for the selected attributes. $Z$ are the constructed clusters, and $T$ are the downstream labels. We find the intersection between the mutual information ($I(Z;T)$) and the negative conditional entropy ($-H(Z|T)$) gives us the best downstream performance.}
\label{fig:attributes}
\vspace{-4mm}
\end{figure}

\vspace{-2mm}
\paragraph{Results.} Table~\ref{tbl:attributes} presents our results. First, we compare different cluster constructions along with Cl-InfoNCE and use the top-1 accuracy on Wider Attribute for discussions. We find the performance grows from low to high when having the clusters as instance ID ($40.2$), 
attributes-determined clusters ($45.5$) to labels ($49.9$). This result suggests that CL-InfoNCE can better bridge the gap with the supervised learned representations by using auxiliary information. Second, we find that using auxiliary information does not always guarantee better performance than not using it. For instance, predicting the attributes-determined clusters using the cross-entropy loss ($39.4$) performs worse than the SimCLR method ($40.2$), which does not utilize the auxiliary information. Hence, how to effectively leverage the auxiliary information is crucial.
Third, we observe the predictive method always performs worse than the contrastive method under the weakly supervised setting. For example, on UT-Zappos50K, although predicting the labels using the cross-entropy loss ($89.2$) performs at par with SupCon ($89.0$), predicting attributes-determined clusters using the cross-entropy loss ($82.7$) performs worse than attributes-determined clusters + Cl-InfoNCE ($84.6$). This result implies that the contrastive method (e.g., Cl-InfoNCE) can generally be applied across various supervision levels.

To better understand the effect of the hyper-parameter $k$ for constructing the attributes-determined clusters, we study the information-theoretical metrics between $Z$ and $T$ and report in Figure~\ref{fig:attributes}. First, as $k$ increases, the mutual information $I(Z;T)$ increases but the conditional entropy $H(Z|T)$ also increases. Hence, although considering more attributes leads to the clusters that are more correlated to the downstream labels, the clusters may also contain more downstream-irrelevant information. This is in accord with our second observation that, as $k$ increases, the downstream performance first increases then decreases. Therefore, we only need a partial set of the most informative attributes (those with high entropy) to determine the clusters. Our last observation is that the best performing clusters happen at the intersection between $I(Z;T)$ and negative $H(Z|T)$. This observation helps us study the trade-off between $I(Z;T)$ and $H(Z|T)$ and suggests that the clusters, when used for Cl-InfoNCE, having the highest $I(Z;T)-H(Z|T)$ could achieve the best performance. 

\vspace{-1mm}
\subsection{Experiment III: Data-Hierarchy-Determined Clusters + Cl-InfoNCE}
\label{subsec:exp_with_auxi_hier}
\vspace{-1mm}

The experimental setup and the comparing baselines are similar to Section~\ref{subsec:exp_with_auxi_attr}, but now we consider the WordNet~\citep{miller1995wordnet} hierarchy as the auxiliary information. As discussed in Section~\ref{subsec:cluster_construct} and Figure~\ref{fig:cluster_construction}, we construct the clusters $Z$ such that the data within a cluster have the same parent node in the level $l$ in the data's WordNet tree hierarchy. $l$ is the hyper-parameter. 

\begin{figure}[t!] 
\vspace{-5mm}
\centering
\includegraphics[width=0.95\textwidth]{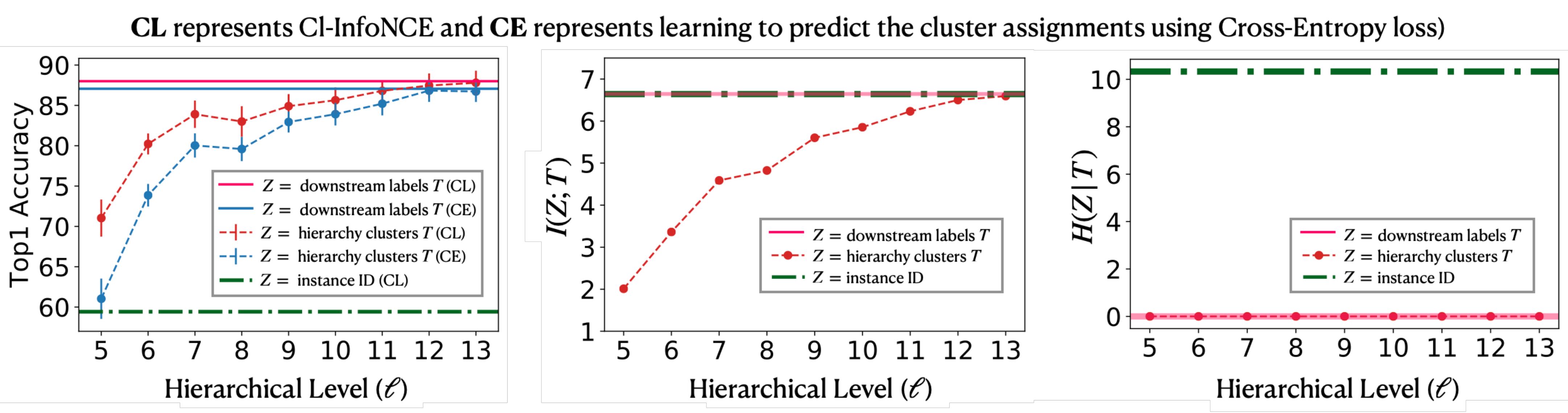}
\vspace{-2mm}
\caption{\small 
Experimental results on ImageNet-100 for Cl-InfoNCE under supervised (clusters $Z =$ downstream labels $T$), weakly supervised ($Z =$ hierarchy clusters) and conventional self-supervised ($Z =$ instance ID) setting. 
We also consider the baseline - learning to predict the clustering assignment using the cross-entropy loss. 
Note that we construct the clusters such that the data within a cluster have the same parent node in the level $\ell$ in the data's WordNet tree hierarchy. Under this construction, the root node is of the level $1$, and the downstream labels are of the level $14$. $I(Z;T)$ is the mutual information, and $H(Z|T)$ is the conditional entropy.}
\vspace{-3mm}
\label{fig:hierarchy}
\end{figure}

\vspace{-1mm}
\paragraph{Results.} Figure~\ref{fig:hierarchy} presents our results. First, we look at the leftmost plot, and we have several similar observations when having the data attributes as the auxiliary information. One of them is that the contrastive method consistently outperforms the predictive method. Another of them is that the weakly supervised representations better close the gap with the supervised representations. Second, as discussed in Section~\ref{subsec:cluster_construct}, the WordNet data hierarchy clusters can be regarded as the coarse labels of the data. Hence, when increasing the hierarchy level $l$, we can observe the performance improvement (see the leftmost plot) and the increasing mutual information $I(Z;T)$ (see the middle plot) between the clusters $Z$ and the labels $T$. Note that $H(Z|T)$ remains zero (see the rightmost plot) since the coarse labels (the intermediate nodes) can be determined by the downstream labels (the leaf nodes) under the tree hierarchy structure. Third, we discuss the conventional self-supervised setting with the special case when $Z=$ instanced ID. $Z$ as the instance ID has the highest $I(Z;T)$ (see the middle plot) but also the highest $H(Z|T)$ (see the rightmost plot). And we observe that the conventional self-supervised representations perform the worse (see the leftmost plot). We conclude that, when using cluster-based representation learning approaches, we shall not rely purely on the mutual information between the data clusters and the downstream labels to determine the goodness of the learned representations. We shall also take the redundant information in the clusters into account.

\vspace{-1mm}
\section{Conclusion and Discussions}
\label{sec:conclu}
\vspace{-1mm}

In this paper, we present to integrate auxiliary information of data into the self-supervised learning process. We first construct data clusters according to auxiliary information. Then, we introduce the clustering InfoNCE (Cl-InfoNCE) objective to leverage the built clusters. Our method brings the performance closer to the supervised learned representations compared to the conventional self-supervised learning approaches. Moreover, even without auxiliary information, Cl-InfoNCE can work with unsupervised K-means clusters as a strong method under the conventional self-supervised learning setting. We believe this work sheds light on the advantage of exploiting 1) noisy but cheap-to-collect sources of information in the wild and 2) data structure information for learning better representations.  

\vspace{-1mm}
\paragraph{Limitations.} Our approach requires determining data clusters from auxiliary information. In our paper, we present different data cluster construction methods for discrete attributes and data hierarchy information. Nonetheless, some types of auxiliary information may be highly unstructured. And determining the clusters according to such auxiliary information may require additional effort. For instance, if having continuous attributes as auxiliary information, binning or quantization cannot be avoided when constructing the clusters.

\vspace{-1mm}
\paragraph{Negative Social Impacts.} Certain auxiliary information may contain private information. For example, in medical applications, physical conditions as auxiliary information may reveal a person's identity. Therefore, we should be careful in choosing auxiliary information for privacy concerns.

{
\small
\bibliography{ref}
\bibliographystyle{plainnat}
}

\appendix

\section{Theoretical Analysis}

In this section, we provide theoretical analysis on the presented Cl-InfoNCE objective. We recall the proposition of Cl-InfoNCE and our presented theorem:

\begin{proposition}[Clustering-based InfoNCE (Cl-InfoNCE), restating Proposition 3.1 in the main text] 
\begin{equation*}
{\rm Cl-InfoNCE}:=\underset{f}{\rm sup}\,\,\mathbb{E}_{(x_i, y_i)\sim {\mathbb{E}_{z\sim P_Z}\big[P_{X|z}P_{Y|z}\big]}^{\otimes n}}\Big[ {\rm log}\,\frac{e^{f(x_i, y_i)}}{\frac{1}{n}\sum_{j=1}^n e^{f(x_i, y_j)}}\Big],
\end{equation*}
\label{prop:cl-infonce-2}
\end{proposition}
\begin{theorem}[informal, Cl-InfoNCE maximization learns to include the clustering information, restating Theorem 3.2 in the main text]
\begin{equation*}
\begin{split}
    & {\rm Cl-InfoNCE} \leq D_{\rm KL}\,\Big( \mathbb{E}_{P_Z}\big[P_{X|Z}P_{Y|Z}\big] \,\|\, P_{X}P_{Y} \Big) \leq H(Z) \\
{\rm and}\,\,& {\rm the\,\,equality\,\,holds\,\,only\,\,when \,\,} H(Z|X) = H(Z|Y) = 0.
\end{split}
\end{equation*}
\label{theo:max_resulting_repre_2}
\end{theorem}

Our goal is to prove Theorem~\ref{theo:max_resulting_repre_2}. For a better presentation flow, we split the proof into three parts:
\begin{itemize}
    \item Proving ${\rm Cl-InfoNCE} \leq D_{\rm KL}\,\Big( \mathbb{E}_{P_Z}\big[P_{X|Z}P_{Y|Z}\big] \,\|\, P_{X}P_{Y} \Big)$ in Section~\ref{subsec:proof_a}
    \item Proving $D_{\rm KL}\,\Big( \mathbb{E}_{P_Z}\big[P_{X|Z}P_{Y|Z}\big] \,\|\, P_{X}P_{Y} \Big) \leq H(Z)$ in Section~\ref{subsec:proof_b}
    \item Proving ${\rm Cl-InfoNCE} {\rm \,\,maximizes\,\,at\,\,} H(Z) {\rm \,\,when \,\,} H(Z|X) = H(Z|Y) = 0$ in Section~\ref{subsec:proof_c}
\end{itemize}

\subsection{Part I - Proving ${\rm Cl-InfoNCE} \leq D_{\rm KL}\,\Big( \mathbb{E}_{P_Z}\big[P_{X|Z}P_{Y|Z}\big] \,\|\, P_{X}P_{Y} \Big)$}
\label{subsec:proof_a}
The proof requires the following lemma. 

\begin{lemma}[Theorem 1 by~\citet{song2020multi}] Let $\mathcal{X}$ and $\mathcal{Y}$ be the sample spaces for $X$ and $Y$, $f$ be any function: $(\mathcal{X} \times \mathcal{Y}) \rightarrow \mathbb{R}$, and $\mathcal{P}$ and $\mathcal{Q}$ be the probability measures on $\mathcal{X} \times \mathcal{Y}$. Then,
$$
\underset{f}{\rm sup}\,\,\mathbb{E}_{(x, y_1)\sim \mathcal{P}, (x, y_{2:n})\sim \mathcal{Q}^{\otimes (n-1)}}\Big[ {\rm log}\,\frac{e^{f(x, y_1)}}{\frac{1}{n}\sum_{j=1}^n e^{f(x, y_j)}}\Big] \leq D_{\rm KL} \Big( \mathcal{P} \,\|\, \mathcal{Q} \Big).
$$
\label{lemm:infonce_like}
\end{lemma}

Now, we are ready to prove the following lemma: 
\begin{lemma}[Proof Part I]
$
    {\rm Cl-InfoNCE}:=\underset{f}{\rm sup}\,\,\mathbb{E}_{(x_i, y_i)\sim {\mathbb{E}_{z\sim P_Z}\big[P_{X|z}P_{Y|z}\big]}^{\otimes n}}\Big[ {\rm log}\,\frac{e^{f(x_i, y_i)}}{\frac{1}{n}\sum_{j=1}^n e^{f(x_i, y_j)}}\Big] \leq D_{\rm KL}\,\Big( \mathbb{E}_{P_Z}\big[P_{X|Z}P_{Y|Z}\big] \,\|\, P_{X}P_{Y} \Big).
$
\begin{proof}
By defining $\mathcal{P} = \mathbb{E}_{P_Z}\big[P_{X|Z}P_{Y|Z}\big]$ and $\mathcal{Q} = P_XP_Y$, we have 
$$
\mathbb{E}_{(x, y_1)\sim \mathcal{P}, (x, y_{2:n})\sim \mathcal{Q}^{\otimes (n-1)}}\Big[ {\rm log}\,\frac{e^{f(x, y_1)}}{\frac{1}{n}\sum_{j=1}^n e^{f(x, y_j)}}\Big] = \mathbb{E}_{(x_i, y_i)\sim {\mathbb{E}_{z\sim P_Z}\big[P_{X|z}P_{Y|z}\big]}^{\otimes n}}\Big[ {\rm log}\,\frac{e^{f(x_i, y_i)}}{\frac{1}{n}\sum_{j=1}^n e^{f(x_i, y_j)}}\Big].
$$
Plug in this result into Lemma~\ref{lemm:infonce_like} and we conclude the proof.
\end{proof}
\label{lemm:part_a}
\end{lemma}

\subsection{Part II - Proving $D_{\rm KL}\,\Big( \mathbb{E}_{P_Z}\big[P_{X|Z}P_{Y|Z}\big] \,\|\, P_{X}P_{Y} \Big) \leq H(Z)$}
\label{subsec:proof_b}

The proof requires the following lemma:
\begin{lemma}
$
D_{\rm KL}\,\Big( \mathbb{E}_{P_Z}\big[P_{X|Z}P_{Y|Z}\big] \,\|\, P_{X}P_{Y} \Big) \leq {\rm min}\,\Big\{{\rm MI}(Z;X), {\rm MI}(Z;Y)\Big\}.
$

\begin{proof}
\begin{equation*}
\begin{split}
    & {\rm MI}(Z;X) - D_{\rm KL}\,\Big( \mathbb{E}_{P_Z}\big[P_{X|Z}P_{Y|Z}\big] \,\|\, P_{X}P_{Y} \Big) \\
    = & \int_z p(z) \int_x p(x|z) \,\,{\rm log}\,\,\frac{p(x|z)}{p(x)} {\rm d}x {\rm d}z -  \int_z p(z) \int_x p(x|z) \int_y p(y|z) \,\,{\rm log}\,\,\frac{\int_{z'}p(z')p(x|z')p(y|z'){\rm d}z'}{p(x)p(y)} {\rm d}x {\rm d}y {\rm d}z \\
    = & \int_z p(z) \int_x p(x|z) \,\,{\rm log}\,\,\frac{p(x|z)}{p(x)} {\rm d}x {\rm d}z -  \int_z p(z) \int_x p(x|z) \int_y p(y|z) \,\,{\rm log}\,\,\frac{\int_{z'}p(z'|y)p(x|z'){\rm d}z'}{p(x)} {\rm d}x {\rm d}y {\rm d}z \\
    = & \int_z p(z) \int_x p(x|z) \int_y p(y|z) \,\,{\rm log}\,\,\frac{p(x|z)}{\int_{z'}p(z'|y)p(x|z'){\rm d}z'}{\rm d}x {\rm d}y {\rm d}z \\
    = & - \int_z p(z) \int_x p(x|z) \int_y p(y|z) \,\,{\rm log}\,\,\frac{\int_{z'}p(z'|y)p(x|z'){\rm d}z'}{p(x|z)}{\rm d}x {\rm d}y {\rm d}z \\
    \geq & - \int_z p(z) \int_x p(x|z) \int_y p(y|z) \,\,\Bigg(\frac{\int_{z'}p(z'|y)p(x|z'){\rm d}z'}{p(x|z)} - 1\Bigg){\rm d}x {\rm d}y {\rm d}z \,\,\Big(\because {\rm log}\,t \leq t-1  \Big) \\
    = &\,\, 0.
\end{split}
\end{equation*}
Hence, ${\rm MI}(Z;X) \geq D_{\rm KL}\,\Big( \mathbb{E}_{P_Z}\big[P_{X|Z}P_{Y|Z}\big] \,\|\, P_{X}P_{Y} \Big)$. Likewise, ${\rm MI}(Z;Y) \geq D_{\rm KL}\,\Big( \mathbb{E}_{P_Z}\big[P_{X|Z}P_{Y|Z}\big] \,\|\, P_{X}P_{Y} \Big)$. We complete the proof by combining the two results.
\end{proof}
\label{lemm:less_than_mi}
\end{lemma}

Now, we are ready to prove the following lemma:
\begin{lemma}[Proof Part II]
$D_{\rm KL}\,\Big( \mathbb{E}_{P_Z}\big[P_{X|Z}P_{Y|Z}\big] \,\|\, P_{X}P_{Y} \Big) \leq H(Z).$
\begin{proof}
Combining Lemma~\ref{lemm:less_than_mi} and the fact that ${\rm min}\,\Big\{{\rm MI}(Z;X), {\rm MI}(Z;Y)\Big\} \leq H(Z)$, we complete the proof. Note that we consider $Z$ as the clustering assignment, which is discrete but not continuous. And the inequality holds for the discrete $Z$, but may not hold for the continuous $Z$.
\end{proof}
\label{lemm:part_b}
\end{lemma}

\subsection{Part III - Proving ${\rm Cl-InfoNCE} {\rm \,\,maximizes\,\,at\,\,} H(Z) {\rm \,\,when \,\,} H(Z|X) = H(Z|Y) = 0$}
\label{subsec:proof_c}

We directly provide the following lemma:
\begin{lemma}[Proof Part III]
${\rm Cl-InfoNCE} {\rm \,\,max.\,\,at\,\,} H(Z) {\rm \,\,when \,\,} H(Z|X) = H(Z|Y) = 0.$
\begin{proof}
When $H(Z|Y) = 0$, $p(Z|Y=y)$ is Dirac. The objective 
\begin{equation*}
\begin{split}
    & D_{\rm KL}\,\Big( \mathbb{E}_{P_Z}\big[P_{X|Z}P_{Y|Z}\big] \,\|\, P_{X}P_{Y} \Big) \\
    = &  \int_z p(z) \int_x p(x|z) \int_y p(y|z) \,\,{\rm log}\,\,\frac{\int_{z'}p(z')p(x|z')p(y|z'){\rm d}z'}{p(x)p(y)} {\rm d}x {\rm d}y {\rm d}z \\
    = & \int_z p(z) \int_x p(x|z) \int_y p(y|z) \,\,{\rm log}\,\,\frac{\int_{z'}p(z'|y)p(x|z'){\rm d}z'}{p(x)} {\rm d}x {\rm d}y {\rm d}z \\
    = & \int_z p(z) \int_x p(x|z) \int_y p(y|z) \,\,{\rm log}\,\,\frac{\int_{z'}p(z')p(x|z')p(y|z'){\rm d}z'}{p(x)p(y)} {\rm d}x {\rm d}y {\rm d}z \\
    = & \int_z p(z) \int_x p(x|z) \int_y p(y|z) \,\,{\rm log}\,\,\frac{p(x|z)}{p(x)} {\rm d}x {\rm d}y {\rm d}z = {\rm MI}\Big( Z;X\Big).
\end{split}
\end{equation*}
The second-last equality comes with the fact that: when $p(Z|Y=y)$ is Dirac, $p(z'|y) = 1 \,\,\forall z' = z$ and $p(z'|y) = 0 \,\,\forall z' \neq z$. Combining with the fact that ${\rm MI}\Big( Z;X\Big) = H(Z)$ when $H(Z|X)=0$, we know 
$D_{\rm KL}\,\Big( \mathbb{E}_{P_Z}\big[P_{X|Z}P_{Y|Z}\big] \,\|\, P_{X}P_{Y} \Big) = H(Z) $ when $H(Z|X)=H(Z|Y)=0$.

Furthermore, by Lemma~\ref{lemm:part_a} and Lemma~\ref{lemm:part_b}, we complete the proof.
\end{proof}
\label{lemm:part_c}
\end{lemma}

\subsection{Bringing Everything Together}

We bring Lemmas~\ref{lemm:part_a},~\ref{lemm:part_b}, and~\ref{lemm:part_c} together and complete the proof of Theorem~\ref{theo:max_resulting_repre_2}.

\section{Algorithms}
In this section, we provide algorithms for our experiments. We consider two sets of the experiments. The first one is K-means clusters + Cl-InfoNCE (see Section 4.3 in the main text), where the clusters involved in Cl-InfoNCE are iteratively obtained via K-means clustering on top of data representations. The second one is auxiliary-information-determined clusters + Cl-InfoNCE (see Section 4.4 and 4.5 in the main text), where the clusters involved in Cl-InfoNCE are pre-determined accordingly to data attributes (see Section 4.4) or data hierarchy information (see Section 4.5).

\paragraph{K-means clusters + Cl-InfoNCE}
We present here the algorithm for K-means clusters + Cl-InfoNCE. At each iteration in our algorithm, we perform K-means Clustering algorithm on top of data representations for obtaining cluster assignments. The cluster assignment will then be used in our Cl-InfoNCE objective.

\begin{algorithm}[H]
\SetAlgoLined
\KwResult{Pretrained Encoder $f_{\theta}(\cdot)$}
 $f_{\theta}(\cdot)\leftarrow \text{Base Encoder Network}$\;
 Aug $(\cdot)\leftarrow$ Obtaining Two Variants of Augmented Data via Augmentation Functions\;
Embedding $\leftarrow$ Gathering data representations by passing data through $f_{\theta}(\cdot)$\;
  Clusters $\leftarrow$\textbf{K-means-clustering}(Embedding)\;
 \For {epoch in 1,2,...,N}{
 \For{batch in 1,2,...,M}{
    data1, data2 $\leftarrow$ Aug(data\_batch)\;
    feature1, feature2 $\leftarrow$ $f_{\theta}$(data1), $f_{\theta}$(data2)\;
    $L_{\text{Cl-infoNCE}}\leftarrow$ Cl-InfoNCE(feature1, feature2, Clusters)\;
    $f_{\theta} \leftarrow f_{\theta} - lr * \frac{\partial}{\partial \theta}L_{\text{Cl-infoNCE}}$\;
 }
 Embedding $\leftarrow$ gather embeddings for all data through $f_{\theta}(\cdot)$\;
  Clusters $\leftarrow$\textbf{K-means-clustering}(Embedding)\;
 }
 \caption{K-means Clusters + Cl-InfoNCE}
\end{algorithm}

\paragraph{Auxiliary information determined clusters + Cl-InfoNCE}
We present the algorithm to combine auxiliary-information-determined clusters with Cl-InfoNCE. We select data attributes or data hierarchy information as the auxiliary information, and we present their clustering determining steps in Section 3.1 in the main text. 

\begin{algorithm}[H]
\SetAlgoLined
\KwResult{Pretrained Encoder $f_{\theta}(\cdot)$}
 $f_{\theta}(\cdot)\leftarrow \text{Base Encoder Network}$\;
 Aug $(\cdot)\leftarrow$ Obtaining Two Variants of Augmented Data via Augmentation Functions\;
 Clusters $\leftarrow$Pre-determining Data Clusters from \textbf{Auxiliary Information}\;
 \For {epoch in 1,2,...,N}{
 \For{batch in 1,2,...,M}{
    data1, data2 $\leftarrow$ Aug(data\_batch)\;
    feature1, feature2 $\leftarrow$ $f_{\theta}$(data1), $f_{\theta}$(data2)\;
    $L_{\text{Cl-infoNCE}}\leftarrow$ Cl-InfoNCE(feature1, feature2, Clusters)\;
    $f_{\theta} \leftarrow f_{\theta} - lr * \frac{\partial}{\partial \theta}L_{\text{Cl-infoNCE}}$\;
 }
 }
 \caption{Pre-Determined Clusters + Cl-InfoNCE}
\end{algorithm}

\section{Experimental details}
The following content describes our experiments settings in details. For reference, our code is avaiable at \url{https://anonymous.4open.science/r/Cl-InfoNCE-02AB/README.md}.

\subsection{UT-Zappos50K}
The following section describes the experiments we performed on UT-Zappos50K dataset in Section 4 in the main text.
\paragraph{Accessiblity}
The dataset is attributed to \citep{yu2014fine} and available at the link: \url{http://vision.cs.utexas.edu/projects/finegrained/utzap50k}. The dataset is for non-commercial use only.

\paragraph{Data Processing}
The dataset contains images of shoe from Zappos.com. We rescale the images to $32\times 32$. The official dataset has 4 large categories following 21 sub-categories. We utilize the 21 subcategories for all our classification tasks. The dataset comes with 7 attributes as auxiliary information. We binarize the 7 discrete attributes into 126 binary attributes. We rank the binarized attributes based on their entropy and use the top-$k$ binary attributes to form clusters. Note that different $k$ result in different data clusters (see Figure 5 (a) in the main text).

\textit{Training and Test Split}: We randomly split train-validation images by $7:3$ ratio,  resulting in $35,017$ train data and $15,008$ validation dataset. 

\paragraph{Network Design}
We use ResNet-50 architecture to serve as a backbone for encoder. To compensate the 32x32 image size, we change the first 7x7 2D convolution to 3x3 2D convolution and remove the first max pooling layer in the normal ResNet-50 (See code for detail). This allows finer grain of information processing. After using the modified ResNet-50 as encoder, we include a 2048-2048-128 Multi-Layer Perceptron (MLP) as the projection head \Big(i.e., $g(\cdot)$ in $f(\cdot, \cdot)$ equation (1) in the main text\Big) for Cl-InfoNCE. During evaluation, we discard the projection head and train a linear layer on top of the encoder's output. For both K-means clusters + Cl-InfoNCE and auxiliary-information-determined clusters + Cl-InfoNCE, we adopt the same network architecture, including the same encoder, the same MLP projection head and the same linear evaluation protocol. In the K-means + Cl-InfoNCE settings, the number of the K-means clusters is $1,000$. Kmeans clustering is performed every epoch during training. We find performing Kmeans for every epoch benefits the performance. For fair comparsion, we use the same network architecture and cluster number for PCL.

\paragraph{Optimization}
We choose SGD with momentum of $0.95$ for optimizer with a weight decay of $0.0001$ to prevent network over-fitting. To allow stable training, we employ a linear warm-up and cosine decay scheduler for learning rate. For experiments shown in Figure 5 (a) in the main text, the learning rate is set to be $0.17$ and the temperature is chosen to be $0.07$ in Cl-InfoNCE. And for experiments shown in Figure 4 in the main text, learning rate is set to be $0.1$ and the temperature is chosen to be $0.1$ in Cl-InfoNCE. 

\paragraph{Computational Resource}
We conduct experiments on machines with 4 NVIDIA Tesla P100. It takes about 16 hours to run 1000 epochs of training with batch size 128 for both auxiliary information aided and unsupervised Cl-InfoNCE.

\subsection{Wider Attributes}
The following section describes the experiments we performed on Wider Attributes dataset in Section 4 in the main text.
\paragraph{Accessiblity}
The dataset is credited to \citep{li2016human} and can be downloaded from the link: \url{http://mmlab.ie.cuhk.edu.hk/projects/WIDERAttribute.html}. The dataset is for public and non-commercial usage.

\paragraph{Data Processing}
The dataset contains $13,789$ images with multiple semantic bounding boxes attached to each image. Each bounding is annotated with $14$ binary attributes, and different bounding boxes in an image may have different attributes. Here, we perform the OR operation among the attributes in the bounding boxes in an image. Hence, each image is linked to $14$ binary attributes. We rank the 14 attributes by their entropy and use the top-$k$ of them when performing experiments in Figure 5 (b) in the main text. We consider a classification task consisting of $30$ scene categories. 

\textit{Training and Test Split}: The dataset comes with its training, validation, and test split. Due to a small number of data, we combine the original training and validation set as our training set and use the original test set as our validation set. The resulting training set contains $6,871$ images and the validation set contains $6,918$ images.

\paragraph{Computational Resource}
To speed up computation, on Wider Attribute dataset we use a batch size of $40$, resulting in 16-hour  computation in a single NVIDIA Tesla P100 GPU for $1,000$ epochs training.  

\paragraph{Network Design and Optimization}
We use ResNet-50 architecture as an encoder for Wider Attributed dataset. We choose 2048-2048-128 MLP as the projection head \Big(i.e., $g(\cdot)$ in $f(\cdot, \cdot)$ equation (1) in the main text\Big) for Cl-InfoNCE. The MLP projection head is discarded during the linear evaluation protocol. Particularly, during the linear evaluation protocol, the encoder is frozen and a linear layer on top of the encoder is fine-tuned with downstream labels.  For Kmeans + Cl-InfoNCE and Auxiliary information + Cl-InfoNCE, we consider the same architectures for the encoder, the MLP head and the linear evaluation classifier. For K-means + Cl-InfoNCE, we consider $1,000$ K-means clusters. For fair comparsion, the same network architecture and cluster number is used for experiments with PCL.

For Optimization, we use SGD with momentum of $0.95$. Additionally, $0.0001$ weight decay is adopted in the network to prevent over-fitting. We use a learning rate of $0.1$ and temperature of $0.1$ in Cl-InfoNCE for all experiments. A linear warm-up following a cosine decay is used for the learning rate scheduling, providing a more stable learning process. 

\subsection{CUB-200-2011}
The following section describes the experiments we performed on CUB-200-2011 dataset in Section 4 in the main text.
\paragraph{Accessiblity}
CUB-200-2011 is created by \citet{wah2011caltech} and is a fine-grained dataset for bird species. It can be downloaded from the link: \url{http://www.vision.caltech.edu/visipedia/CUB-200-2011.html}. The usage is restricted to non-commercial research and educational purposes. 

\paragraph{Data Processing}
The original dataset contains $200$ birds categories over $11,788$ images with $312$ binary attributes attached to each image. We utilize those attributes and rank them based on their entropy. In Figure 5 (c), we use the top-$k$ of those attributes to constrcut clusters with which we perform in Cl-InfoNCE. The image is rescaled to $224\times 224$.

\textit{Train Test Split}:
We follow the original train-validation split, resulting in $5,994$ train images and $5,794$ validation images. 

\paragraph{Computational Resource}
It takes about 8 hours to train for 1000 epochs with 128 batch size on 4 NVIDIA Tesla P100 GPUs. 

\paragraph{Network Design and Optimization}
We choose ResNet-50 for CUB-200-2011 as the encoder. After extracting features from the encoder, a 2048-2048-128 MLP projection head \Big(i.e., $g(\cdot)$ in $f(\cdot, \cdot)$ equation (1) in the main text\Big) is used for Cl-InfoNCE. During the linear evaluation protocal, the MLP projection head is removed and the features extracted from the pre-trained encoder is fed into a linear classifier layer. The linear classifier layer is fine-tuned with the downstream labels. The network architectures remain the same for both K-means clusters + Cl-InfoNCE and auxiliary-information-determined clusters + Cl-InfoNCE settings. In the K-means clusters + Cl-InfoNCE settings, we consider $1,000$ K-means clusters. For fair comparsion, the same network architecture and cluster number is used for experiments with PCL.

SGD with momentum of $0.95$ is used during the optimization. We select a linear warm-up following a cosine decay learning rate scheduler. The peak learning rate is chosen to be $0.1$ and the temperature is set to be $0.1$ for both K-means + Cl-InfoNCE and Auxiliary information + Cl-InfoNCE settings.

\subsection{ImageNet-100}
The following section describes the experiments we performed on ImageNet-100 dataset in Section 4 in the main text.
\paragraph{Accessibility}
This dataset is a subset of ImageNet-1K dataset, which comes from the ImageNet Large Scale Visual Recognition Challenge (ILSVRC) 2012-2017 \citep{russakovsky2015imagenet}. ILSVRC is for non-commercial research and educational purposes and we refer to the ImageNet official site for more information: \url{https://www.image-net.org/download.php}.

\paragraph{Data Processing}
In the Section 4.5 in the main text, we select $100$ classes from ImageNet-1K to conduct experiments (the selected categories can be found in \url{https://anonymous.4open.science/r/Cl-InfoNCE-02AB/data_processing/imagenet100/selected_100_classes.txt}). We also conduct a slight pre-processing (via pruning a small number of edges in the WordNet graph) on the WordNet hierarchy structure to ensure it admits a tree structure. Specifically, each of the selected categories and their ancestors only have one path to the root. We refer the pruning procedure in \url{https://anonymous.4open.science/r/Cl-InfoNCE-02AB/data_processing/imagenet100/hierarchy_processing/imagenet_hierarchy.py} (line 222 to 251).

We cluster data according to their common ancestor in the pruned tree structure and determine the level $l$ of each cluster by the step needed to traverse from root to that node in the pruned tree. Therefore, the larger the $l$, the closer the common ancestor is to the real class labels, hence more accurate clusters will be formed. Particularly, the real class labels is at level $14$. 

\textit{Training and Test Split}: 
Please refer to the following file for the training and validation split.
\begin{itemize}
    \item training: \url{https://anonymous.4open.science/r/Cl-InfoNCE-02AB/data_processing/imagenet100/hier/meta_data_train.csv}
    \item validation: \url{https://anonymous.4open.science/r/Cl-InfoNCE-02AB/data_processing/imagenet100/hier/meta_data_val.csv}
\end{itemize}
The training split contains $128,783$ images and the test split contains $5,000$ images. The images are rescaled to size $224\times 224$.

\paragraph{Computational Resource}
It takes $48$-hour training for $200$ epochs with batch size $128$ using $4$ NVIDIA Tesla P100 machines. All the experiments on ImageNet-100 is trained with the same batch size and number of epochs.  

\paragraph{Network Design and Optimization Hyper-parameters}
We use conventional ResNet-50 as the backbone for the encoder. 2048-2048-128 MLP layer and $l2$ normalization layer is used after the encoder during training and discarded in the linear evaluation protocal. We maintain the same architecture for Kmeans + Cl-InfoNCE and auxiliary information aided Cl-InfoNCE. For Kmeans + Cl-InfoNCE, we choose 2500 as the cluster number. For fair comparsion, the same network architecture and cluster number is used for experiments with PCL. The Optimizer is SGD with $0.95$ momentum. For K-means + Cl-InfoNCE used in Figure 4 in the main text, we use the learning rate of $0.03$ and the temperature of $0.2$. We use the learning rate of $0.1$ and temperature of $0.1$ for auxiliary information + Cl-InfoNCE in Figure 6 in the main text. A linear warm-up and cosine decay is used for the learning rate scheduling. To stablize the training and reduce overfitting, we adopt $0.0001$ weight decay for the encoder network.

\section{Comparisons with Swapping Clustering Assignments between Views}
In this section, we provide additional comparisons between Kmeans + Cl-InfoNCE and Swapping Clustering Assignments between Views (SwAV)~\citep{caron2020unsupervised}. The experiment is performed on ImageNet-100 dataset. SwAV is a recent art for clustering-based self-supervised approach. In particular, SwAV adopts Sinkhorn algorithm~\citep{cuturi2013sinkhorn} to determine the data clustering assignments for a batch of data samples, and SwAV also ensures augmented views of samples will have the same clustering assignments. We present the results in Table~\ref{tab:swav}, where we see SwAV has similar performance with the Prototypical Contrastive Learning method~\citep{li2020prototypical} and has worse performance than our method (i.e., K-means +Cl-InfoNCE).

\begin{table}[h]
\centering
\scalebox{0.8}{
\begin{tabular}{cc}
\toprule
Method              & Top-1 Accuracy (\%) \\ \midrule \midrule  \multicolumn{2}{c}{\it Non-clustering-based Self-supervised Approaches}  \\ \midrule \midrule
SimCLR~\citep{chen2020simple}       & 58.2$\pm$1.7   \\ [1mm]
MoCo~\citep{he2020momentum}      & 59.4$\pm$1.6             \\ \midrule \midrule \multicolumn{2}{c}{\it Clustering-based Self-supervised Approaches (\# of clusters = $2.5$K)}  \\ \midrule \midrule
SwAV~\citep{caron2020unsupervised}   & 68.5$\pm$1.0           \\ [1mm]
PCL~\citep{li2020prototypical}      & 68.9$\pm$0.7           \\ [1mm]
K-means + Cl-InfoNCE (ours)      & \textbf{77.9$\pm$0.7}               \\ \bottomrule
\end{tabular}
}
\vspace{1mm}
\caption{Additional Comparsion with SwAV~\citep{caron2020unsupervised} showing its similar performance as PCL on ImageNet-100 dataset.}
\label{tab:swav}
\end{table}

\section{Preliminary results on ImageNet-1K with Cl-InfoNCE}
We have performed experiments on ImageNet-100 dataset, which is a subset of the ImageNet-1K dataset \citep{russakovsky2015imagenet}. We use the batch size of $1,024$ for all the methods and consider $100$ training epochs. We present the comparisons among Supervised Contrastive Learning~\citep{khosla2020supervised}, our method (i.e., WordNet-hierarchy-information-determined clusters + Cl-InfoNCE), and SimCLR~\citep{chen2020simple}. We select the level-$12$ nodes in the WordNet tree hierarchy structures as our hierarchy-determined clusters for Cl-InfoNCE. We report the results in Table~\ref{tab:imagenet-1K}. We find that our method (i.e., hierarchy-determined clusters + Cl-InfoNCE) performs in between the supervised representations and conventional self-supervised representations.


\begin{table}[h]
\centering
\resizebox{0.7\textwidth}{!}{%
\begin{tabular}{cc}
\toprule
Method     & Top-1 Accuracy (\%) \\ \midrule \midrule \multicolumn{2}{c}{\it Supervised Representation Learning ($Z=$ downstream labels $T$)} \\ \midrule \midrule
SupCon~\citep{khosla2020supervised}     & 76.1$\pm$1.7              \\ \midrule \midrule
\multicolumn{2}{c}{\it Weakly Supervised Representation Learning ($Z=$ level $12$ WordNet hierarchy labels)} \\ \midrule \midrule 
Hierarchy-Clusters + Cl-InfoNCE (ours)    & 67.9$\pm$1.5              \\ \midrule \midrule
\multicolumn{2}{c}{\it Self-supervised Representation Learning ($Z=$ instance ID)} \\ \midrule \midrule
SimCLR~\citep{chen2020simple}     & 62.9$\pm$1.2              \\ \bottomrule
\end{tabular}%
}
\vspace{1mm}
\caption{Preliminary results for WordNet-hierarchy-determined clusters + Cl-InfoNCE on ImageNet-1K.}
\label{tab:imagenet-1K}
\end{table}

\section{Synthetically Constructed Clusters in Section 4.2 in the Main Text}


In Section 4.2 in the main text, on the UT-Zappos50K dataset, we synthesize clusters $Z$ for various $I(Z;T)$ and $H(Z|T)$ with $T$ being the downstream labels. There are $86$ configurations of $Z$ in total. Note that the configuration process has no access to data's auxiliary information and among the $86$ configurations we consider the special cases for the supervised \big($Z=T$\big) and the unsupervised setting \big($Z=$ instance ID\big). In specific, when $Z=T$, $I(Z;T)$ reaches its maximum at $H(T)$ and $H(Z|T)$ reaches its minimum at $0$; when $Z=$ instance ID, both $I(Z;T)$ \big(to be $H(T)$\big) and $H(Z|T)$ \big(to be $H(\text{instance ID})$\big) reaches their maximum. The code for generating these $86$ configurations can be found in lines 177-299 in \url{https://anonymous.4open.science/r/Cl-InfoNCE-02AB/data_processing/UT-zappos50K/synthetic/generate.py}.

\end{document}